\documentclass[runningheads]{llncs}
\usepackage{xcolor}
\usepackage{cite}
\usepackage{setspace}

\usepackage{algorithm,algpseudocode}
\usepackage{multicol}

\usepackage{graphicx}
\usepackage{amsmath}
\usepackage{amssymb}
\usepackage{enumitem}
\usepackage{booktabs}
\usepackage{multirow}
\usepackage{stackengine}
\usepackage{float}
\usepackage{subcaption}
\usepackage{bbm}
\usepackage[colorlinks=true]{hyperref}

\DeclareMathOperator*{\argmax}{argmax}
\def\eg{\emph{e.g.}}

\begin{document}
\title{Learning When to Say \emph{``I Don't Know"}}
%
%

\author{Nicholas Kashani Motlagh\inst{1}\orcidID{0000-0001-6229-6212} \and
Jim Davis\inst{1} \and\\ Tim Anderson \inst{2} \and Jeremy Gwinnup \inst{2}}
\authorrunning{N. Kashani Motlagh et al.}


%

\institute{Department of Computer Science and Engineering, Ohio State University\\
\email{\{kashanimotlagh.1,davis.1719\}@osu.edu} \and
Air Force Research Laboratory, Wright-Patterson AFB \\
\email{\{timothy.anderson.20,\ jeremy.gwinnup.1\}@us.af.mil}}

\maketitle              
\begin{abstract}
We propose a new Reject Option Classification technique to identify and remove regions of uncertainty in the decision space for a given neural classifier and dataset. Such existing formulations employ a learned rejection (remove)/selection (keep) function and require either a known cost for rejecting examples or strong constraints on the accuracy or coverage of the \emph{selected} examples. We consider an alternative formulation by instead analyzing the complementary \emph{reject} region and employing a validation set to learn per-class softmax thresholds. The goal is to maximize the accuracy of the selected examples subject to a natural randomness allowance on the \emph{rejected} examples (rejecting more incorrect than correct predictions). We provide results showing the benefits of the proposed method over na\"ively thresholding calibrated/uncalibrated softmax scores with 2-D points, imagery, and text classification datasets using state-of-the-art pretrained models. Source code is available at {\tt https://github.com/osu-cvl/learning-idk}.

\end{abstract}

\section{Introduction}
\label{sec:intro}
Neural classifiers have shown impressive performance in diverse applications ranging from spam identification to medical diagnosis to autonomous driving. However, the typical argmax softmax decision function forces these networks to sometimes yield unreliable predictions. For example, the 10-class feature space shown in Fig. \ref{fig:cinic10_tsne} displays many regions of class overlap/confusion. It can be desirable to abstain from accepting predictions within highly confusing regions. These low-confidence decisions could be discarded or potentially given to a more complex model (or human analyst) to resolve. Our task is to identify these confusing regions by learning an auxiliary rejection/selection function on predictions. We show the results of using a particular rejection function (described later) in Fig. \ref{fig:cinic10_tsne_select} and \ref{fig:cinic10_tsne_reject}, where plots are generated for the \emph{selected} and \emph{rejected} examples. Clearly, these plots show much stronger classifiable selected examples while further supporting the confusability of the rejected examples.

One simple solution to detect under-confident predictions is to  threshold the softmax value of the argmax decision. However, na\"ively thresholding at 0.5 (the boundary of being more confidently correct) or some other ad hoc threshold may not be ideal or optimal. Reject Option Classification methods \cite{Chow1970,Tortorella2000,Pietraszek2005,El-Yaniv2010,selective2017,SelectiveNet,Franc2021} aim to address this problem by endowing a classifier with a \emph{learned} rejection threshold to reject (remove) or select (keep) predictions, enabling the model to ``know what it doesn't know."

\newcommand{\rulesep}{\unskip\hfill{\color{black}\vrule}\hfill\ignorespaces}
\begin{figure}[t]
    \centering
    \begin{subfigure}{.3\textwidth}
        \includegraphics[width=\textwidth]{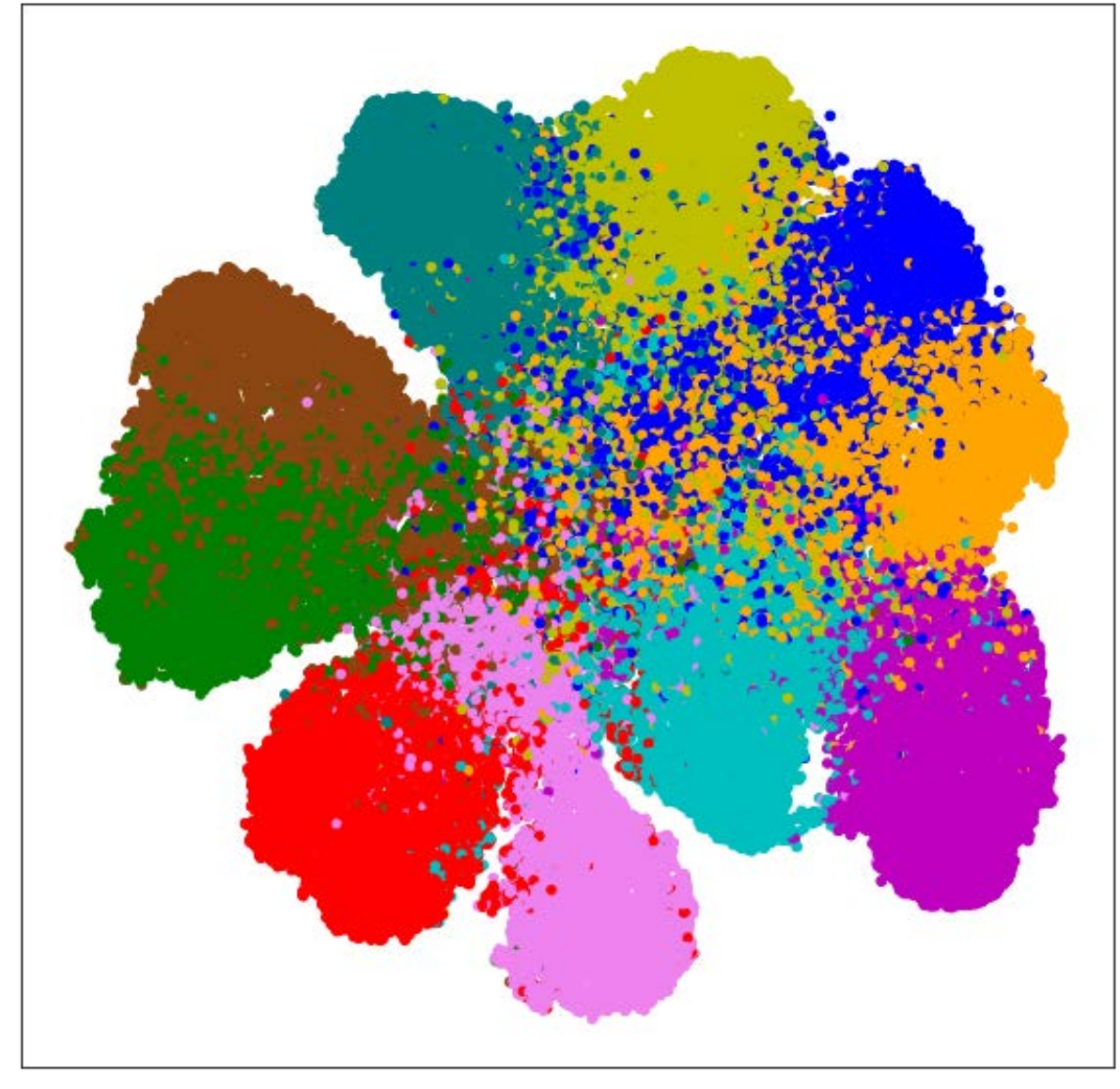}
        \caption{Test set.}
        \label{fig:cinic10_tsne}
    \end{subfigure}%
    \rulesep
    \begin{subfigure}{.3\textwidth}
        \includegraphics[width=\textwidth]{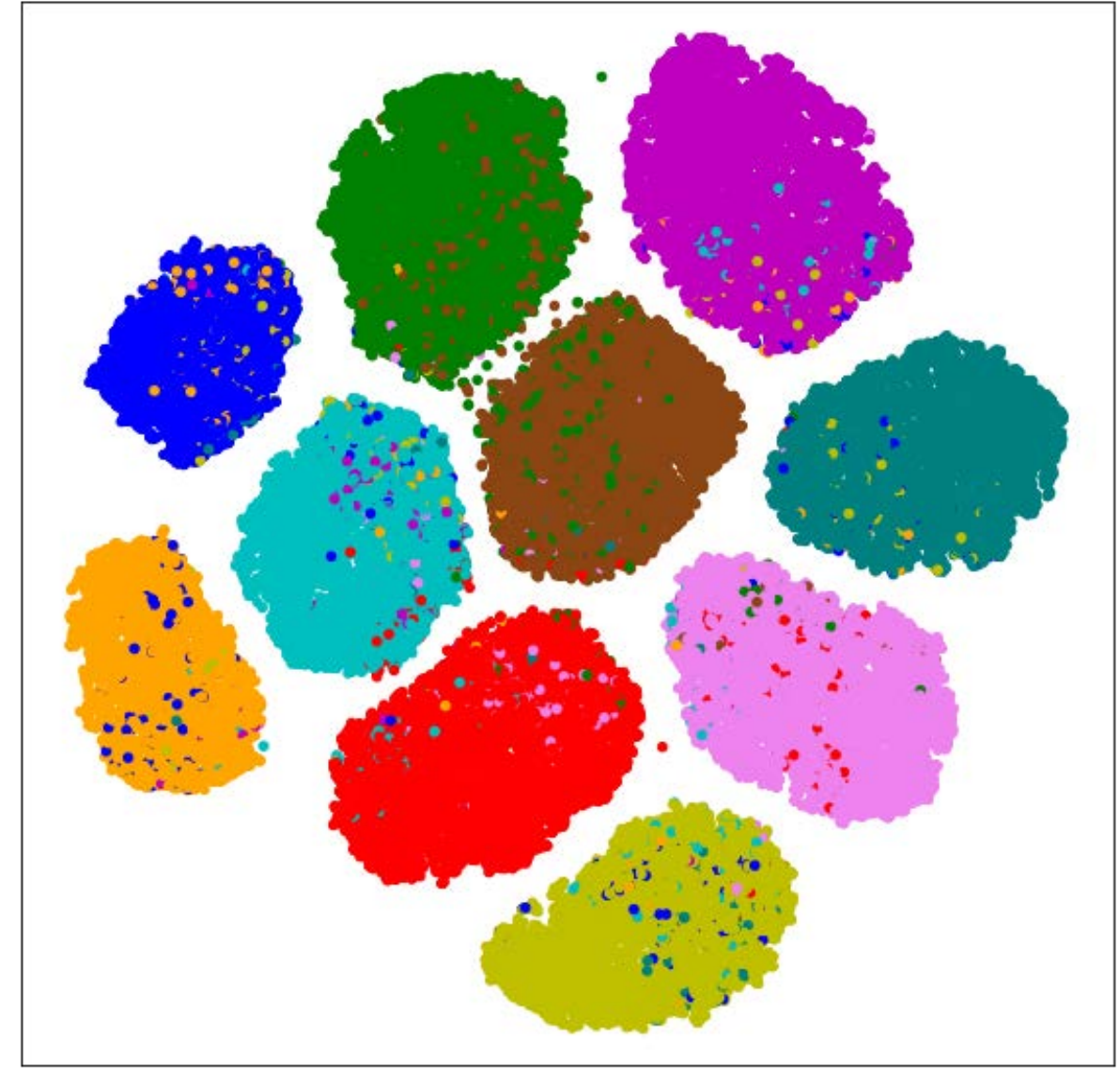}
        \caption{Selected only.}
        \label{fig:cinic10_tsne_select}
    \end{subfigure}%
    \begin{subfigure}{.3\textwidth}
        \includegraphics[width=\textwidth]{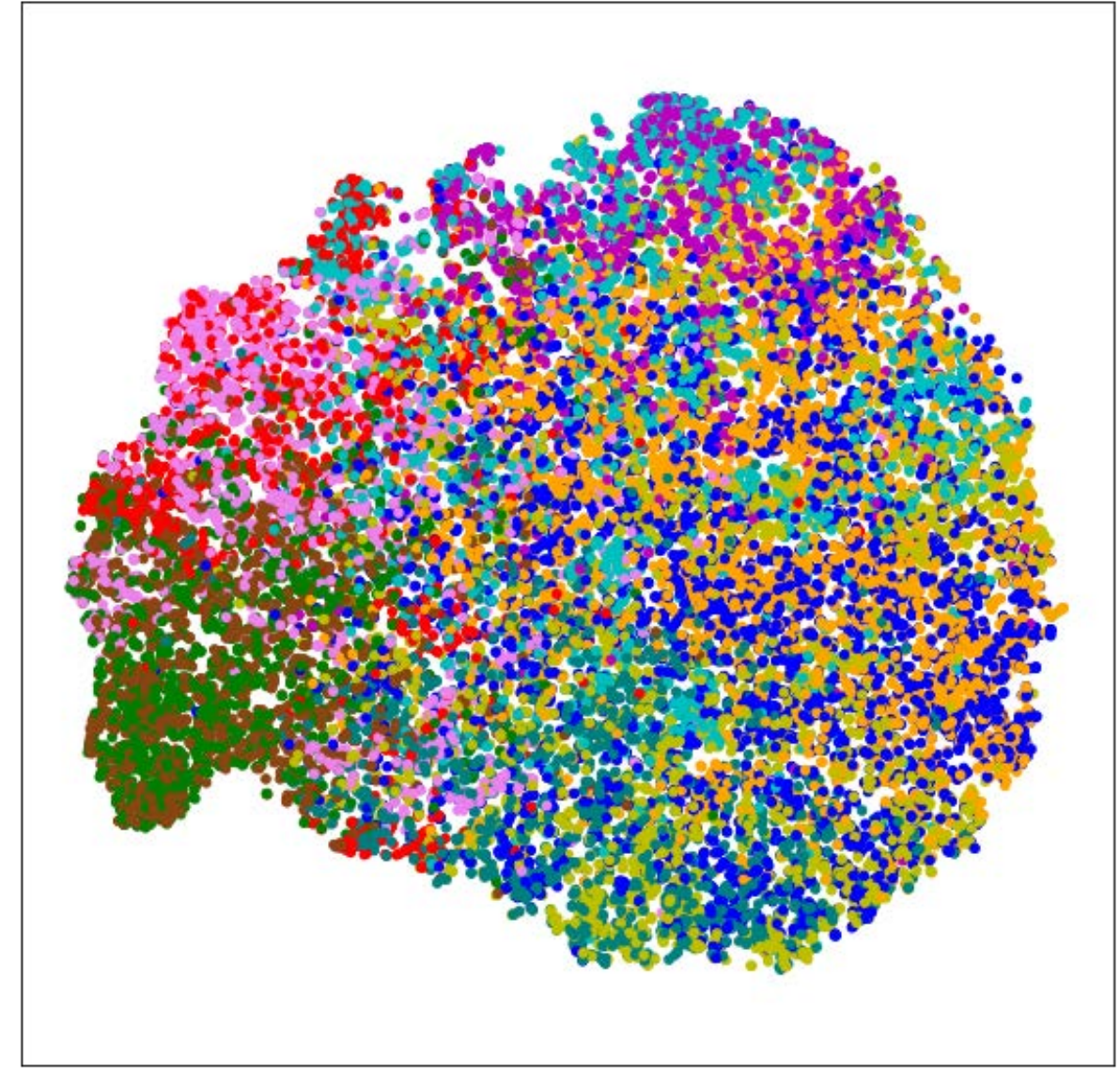}
        \caption{Rejected only.}
        \label{fig:cinic10_tsne_reject}
    \end{subfigure}
    \caption{t-SNE plots of logits from a weakly-trained ResNet20 \cite{resnet} model on CINIC10 \cite{CINIC10} using a) all, b) selected only, and c) rejected only.}
    \label{fig:cinic10_tsne_thresholded}
\end{figure}

 There are three main types of approaches to learning a rejection function. First, the cost-based approach \cite{Chow1970} minimizes an objective that uses a \textit{user-defined} cost for making a rejection. However, applications without a known rejection cost can not leverage this approach. The next type of approach is the bounded-improvement model \cite{Pietraszek2005}, which maximizes coverage (percentage of examples selected) under the constraint that the \emph{select accuracy} (accuracy of selected examples) is lower-bounded by a user-defined amount. For example, a classifier may be required to have a select accuracy of $\geq\!\!95\%$ while trying to maximize the number of selected predictions (coverage). Lastly, the bounded-coverage model \cite{Franc2021} maximizes the select accuracy under the constraint that the \emph{coverage} is lower-bounded by a user-defined amount. For example, it may be necessary to accept $\geq\!\!90\%$ of examples while maximizing the select accuracy.

These previous works focus on user-defined constraints of accuracy and coverage for the selected examples. Here, we present a new Reject Option Classification approach that instead focuses on a natural randomness property desired of \emph{reject} regions. This randomness property is not directly applicable to the complementary select regions and holds across any neural classifier and dataset pairing. Our contributions are summarized as follows:
\begin{enumerate}
    \item Fast post-processing method applicable to any pretrained classifier/dataset.
    \item No user-defined costs/constraints for rejection, select accuracy, or coverage.
    \item Additional approach that reduces computation for large datasets.
\end{enumerate}

\section{Preliminaries}
Let $\mathcal{X}$ be the space of examples, $\mathcal{Y} = \{1,...,c\}$ be a finite label set of $c$ classes, and $P_{\mathcal{X}\mathcal{Y}}$ be the joint data distribution over $\mathcal{X} \times \mathcal{Y}$. Suppose $f$ is a \emph{trained} classifier $f: \mathcal{X} \xrightarrow[]{} [0,1]^c$ with softmax/confidence outputs, and $V_m = \{(x_i,y_i)\}_{i=1}^m$ is a \emph{validation} set of $m$ examples sampled i.i.d. from $P_{\mathcal{X}\mathcal{Y}}$. The empirical accuracy (0-1) of $f$ w.r.t. the validation set $V_m$ is defined as

\begin{equation}
    \label{eqn:acc}
    Acc(f | V_m) \quad=\quad 1- \frac{1}{m}\sum_{i=1}^m \ell(f(x_i),y_i)
\end{equation}

\noindent 
where $\ell(f(x_i),y_i) = \mathbbm{1} [\argmax(f(x_i)) \ne y_i]$ is the 0/1 loss using the indicator function on the argmax decision rule.

A Reject Option Classifier $H_{(f,g_\tau)}$ is defined by a tuple $(f, g_\tau)$ where $f$ is the previously defined trained classifier and $g_\tau:[0,1]^c \xrightarrow[]{} \{0,1\}$ is a rejection function (1:reject, 0:select) with a per-class threshold vector $\tau \in [0,1]^c$. We can write the rejection function for a given example $(x_i,y_i)$ as

\[ g_\tau(f(x_i)) = \mathbbm{1} [\max(f(x_i)) \leq \tau_{\argmax(f(x_i))}]
\]

\noindent Hence, $H_{(f,g_\tau)}$ is

 \[H_{(f,g_\tau)}(x_i) = \begin{cases} 
    \text{reject}, & g_\tau(f(x_i))=1\\
     f(x_i), & g_\tau(f(x_i))=0
      
   \end{cases}
\]

The performance of a Reject Option Classifier can be evaluated using \emph{coverage} ($\phi$) and/or \emph{select accuracy} on the validation set $V_m$. The \emph{coverage} is the proportion of the examples \emph{selected} by $g_\tau$, using

\begin{equation}
    \label{eqn:coverage}
    \phi(H_{(f,g_\tau)}|V_m) \quad=\quad \frac{1}{m}\sum_{i=1}^m (1-g_\tau(f(x_i)))
\end{equation}

\noindent The empirical select accuracy w.r.t. the labeled  validation set $V_m$ (when the coverage is greater than 0) is

\begin{equation}
    \label{eqn:sa}
    SelAcc(H_{(f,g_\tau)} | V_m) \quad=\quad 1 -\frac{\sum_{i=1}^m \ell(f(x_i),y_i) \cdot (1-g_\tau(f(x_i)))}{\sum_{i=1}^m (1-g_\tau(f(x_i)))}
\end{equation}

A trade-off exists between the coverage and select accuracy of a Reject Option Classifier since the only way to \emph{increase} select accuracy is by rejecting more misclassified examples (\emph{decreasing} coverage). Hence, Reject Option Classification problems are typically formulated by either constraining coverage or select accuracy and maximizing the other, as described in the next section.

\section{Related Work}
As previously mentioned, prior approaches to Reject Option Classification employ strong user-defined constraints. In the case where a cost of rejection is available, \cite{Chow1970} provides an optimal strategy when the data distribution $P_{\mathcal{X}\mathcal{Y}}$ is known. In \cite{Tortorella2000}, a model is proposed for binary classification with additional user-defined classification costs using Receiver Operating Characteristic (ROC) analysis. For cases where the cost of rejection is not defined or available, two other constraint-based strategies have been proposed: bounded-improvement and bounded-rejection models.

In \cite{Pietraszek2005}, the \emph{bounded-improvement} model is proposed, where the objective is to maximize the coverage $\phi$ such that the select accuracy has a lower bound of $a^*$, as given by

\begin{equation}
\label{eqnBoundedImprovement}
    \max \phi(H_{(f,g)} | V_m) \quad s.t. \quad {SelAcc(H_{(f,g)} | V_m)} \geq a^*
\end{equation}

\noindent They use ROC analysis to determine optimal decision thresholds in the case of two classes. Furthermore, they assume a classifier can provide output scores (\eg, softmax) proportional to posterior probabilities. In \cite{El-Yaniv2010}, an algorithm is proposed to learn the optimal rejection function when perfect select accuracy is possible for a classifier. The later work of \cite{selective2017} explores the bounded-improvement model in the context of deep neural networks. They propose an algorithm to learn an optimal threshold that statistically guarantees (under a user-defined confidence level) that the \emph{theoretical} select accuracy is greater than a specified target accuracy $a^*$.

Alternatively, the \emph{bounded-coverage} model has the objective of maximizing the select accuracy such that the coverage has a lower bound of $c^*$, as given by

\begin{equation}
\label{eqnBoundedCoverage}
    \max {SelAcc(H_{(f,g)} | V_m)} \quad s.t. \quad \phi(H_{(f,g)} | V_m)\geq c^*
\end{equation}

\noindent In \cite{Franc2021}, the bounded-coverage model is formalized, and a method is provided to obtain uncertainty scores from any black-box classifier. In \cite{SelectiveNet}, a joint training scheme is proposed to simultaneously learn a neural classifier and rejection function that provides the highest accuracy for the desired coverage $c^*$. 

Our proposed method focuses on the \emph{rejected} region rather than the \emph{selected} region. We do not require user-defined costs for rejection nor lower bounds for select accuracy or coverage. Instead, we learn per-class rejection thresholds subject to a natural randomness property desirable of any reject region. Our formulation employs an \emph{upper bound} of randomness (or confusion) expected/desired for a rejection region at a proposed significance level. As multiple viable rejection regions could exist, we maximize in tandem the select accuracy.

\section{Proposed Method}

Consider a \emph{strong} neural classifier (\eg, 99\% accuracy) and an overall rejection threshold chosen to reject all but the single highest scoring prediction. Though the threshold yields 100\% select accuracy (with 1 example), the coverage of the model is far too small (again, just 1 example) for any practical application. Furthermore, the classification accuracy in the remaining reject region would be very high (undesired). Next, consider a \emph{weak} classifier having a large confusion region and a reject threshold set to 0 (reject none, select all). Though the threshold yields 100\% coverage of the examples, the model's select accuracy (of all examples) could be too low to be worth using.

We argue that an \emph{ideal} reject threshold (or set of per-class thresholds) should aim to produce a reject region that has \underline{at most} random-chance classification accuracy. We certainly do \emph{not} want to reject significantly more \emph{correct} than \emph{incorrect} predictions. As multiple viable rejection region sizes may exist, each adhering to at most random-chance classification behavior, the reject region corresponding to the highest accuracy in the complementary accept region should be chosen. Such an approach offers a naturally constrained analysis of reject regions that can be used to filter out indecision areas of \emph{any} neural network and dataset pairing. However, a method must be provided to test a reject region for adherence to the upper-bound randomness requirement. We base our method on a Binomial distribution.

Consider a series of flips of a fair coin with $P(\text{heads})=P(\text{tails})=0.5$ (random chance). Let $Z$ be the number of observed heads after 100 tosses. The expected value of $Z$ is 50, but in reality, the number of heads can deviate from this ideal. The Binomial distribution with $p=0.5$ can be used to calculate the probability of seeing $Z=k$ heads in $n$ trials.
The corresponding Binomial CDF can be used to assess the probability of seeing \emph{at most} $k$ heads

\begin{equation}
\label{eqn:binomial_cdf}
    P(Z \leq k)=BinomCDF(k; n, p)=\sum_{i=0}^k \binom{n}{i} p^{i}(1-p)^{n-i}
\end{equation}

We now shift focus from coin flips to the successes/failures of a classifier. Let $R_\tau = \{(x,y) \in V_m \:\:|\:\: H_{(f,g_\tau)}(x) = \text{reject}\}$ be the set of rejected examples from $V_m$ using threshold $\tau$, where $|R_\tau| = n$. As previously mentioned, we desire a reject region with accuracy $\leq 50\% + \xi$, for some small $\xi$. Note that $\xi$ largely depends on the size of $R_\tau$ (smaller $n$ may allow larger $\xi$ in Binomial probability).

We employ the Binomial CDF model in Eqn. \ref{eqn:binomial_cdf} (with $p=0.5$) along with the \emph{actual} number of classification successes $k^*$ of validation examples (having ground truth) in $R_\tau$ to assess adherence to the desired upper bound random state of $R_\tau$. 
If $BinomCDF(k^*; n, 0.5) = 1-\delta$, this means the probability of getting a \emph{higher} number of correct classifications in the reject region, assuming random behavior, is $\delta$. For example, if $\delta=.05$, there is only a 5\% chance of observing $>\!\!k^*$ correct classifications assuming random chance behavior. Thus, for a given significance level $\delta$, a proposed reject region has too many correct classifications if $BinomCDF(k^*; n, 0.5)>1-\delta$, and therefore must not be accepted. Only when $BinomCDF(k^*; n, 0.5) \leq 1-\delta$ is the region viable. As multiple sizes of a reject region could be deemed viable, we seek to maximize the select accuracy to ensure the highest performing select region from the corresponding set of viable reject regions. To learn the desired per-class rejection thresholds $\tau$, our overall objective, B-CDF$_{\delta}$, for $k^*$ observed successes in the examined reject region is

\begin{equation}
    \label{eqn:proposed}
    \max {SelAcc(H_{(f,g_\tau)} | V_m)} \quad s.t. \quad BinomCDF(k^*; n, 0.5)\leq 1-\delta
\end{equation}

As we are thresholding the softmax scores of the classifier, an important aspect to consider is network calibration, which aims to better align softmax values to true probabilities. Prior Reject Option Classification methods \cite{Chow1970,Tortorella2000,Pietraszek2005,El-Yaniv2010,selective2017,SelectiveNet,Franc2021} did not employ calibration. However, we believe it is an important component to model classification uncertainty properly. Previous work has provided various methods to calibrate networks, including temperature scaling \cite{calibration}. When learning global or per-class thresholds, global temperature scaling is unnecessary (as it is a monotonic operation on the argmax softmax values). Therefore, as promoted in \cite{per-class}, we use \emph{per-class} temperature scaling to better model class-conditional uncertainty before learning per-class rejection thresholds.

The algorithm for our approach is relatively direct and fast to run. We first compute a list of possible rejection thresholds per-class given a validation set. For each class $c$, we identify examples predicted as class $c$. Then, we choose potential thresholds from those examples' calibrated softmax scores. Since we are maximizing select accuracy across viable reject regions, we need only employ thresholds corresponding to the \emph{incorrect} predictions. Next, for a given threshold and class, we extract the corresponding reject region and evaluate the Binomial constraint at a significance level $\delta$. The algorithm chooses the threshold resulting in the highest select accuracy with acceptable reject regions. If multiple thresholds admit the same select accuracy, we choose the smallest threshold (producing the highest coverage). Experiments will demonstrate performance trade-offs across various significance levels $\delta$.

\section{Experiments}
As previous Reject Option Classification methods require strong user/application-defined constraints on select accuracy or coverage, we can not directly compare the bounded-improvement and bounded-coverage models to our method. We instead compare a baseline model (Base) that never rejects any predictions (threshold of 0). Additionally, we compare a na\"ive method that thresholds the softmax values at 0.5. Here, we employ with and without calibration variations, Na\"ive-Cal and Na\"ive-NoCal, respectively. For our proposed Binomial-CDF method (B-CDF$_\delta$), we present results across different significance values $\delta \in \{0.05, 0.1, 0.5, 0.75, 0.95\}$. We use per-class temperature scaling calibration  \cite{per-class} for B-CDF$_\delta$ and the Na\"ive-Cal approaches. 

We evaluate on different modalities: synthetic 2-D point-sets, benchmark image classification datasets, and common text classification datasets. We report and compare various accuracy (ideal, select, reject) and coverage metrics.

We additionally examine the generalization capability of thresholds from validation data to test data. Lastly, we provide an alternative formulation using associated confidence intervals to avoid the computational complexity of the Binomial CDF for very large datasets with vast reject regions.

\subsection{Synthetic Data}
To initially test our approach, we designed 8 multi-class 2-D point datasets with varying amounts of class overlap. We split these datasets into two subsets characterized by class overlap: equal-density and unequal-density. We sampled train-validation-test partitions having 1K-1K-4K examples per-class. We used a test set 4X larger than the training set to better measure the true performance of the learned thresholds. 

For each dataset, we trained a simple neural network consisting of a single hidden layer of 10 nodes followed by a ReLU activation and an output layer consisting of the number of classes (2-4). We trained the network for 50 epochs using a half period cosine learning rate scheduler
(with an initial learning rate of 0.1 and no restarts) and an SGD optimizer with 0.9 momentum. We selected the model from the epoch with the highest validation accuracy. We repeated this process ten times (producing 10 models) using different random seed initializations to provide meaningful statistics on the results \cite{Davis}.

\subsubsection{Equal Density Overlap}\hfill\\
For the first 4 datasets, each class is \emph{uniformly} sampled from a defined rectangular region and positioned such that varying amounts of overlap occur. The top of Table \ref{tab:synth_equal_density} shows the resulting test datasets. Here, the \emph{ideal} reject regions are shown in black and are due to \emph{theoretically} equivalent class densities. To compare results from different reject thresholds, we computed the accuracy of the rejection function (1-reject, 0-select) w.r.t. the ideal reject region, defined as the \emph{Ideal Decision Accuracy} (IDA).

We see the IDA scores of the different approaches in Table \ref{tab:synth_equal_density}. We report the mean IDA score and the standard deviation over the ten trained models for each method and dataset. We also computed one-sided T-tests at the 0.05 significance level \cite{Davis} against the highest mean IDA, and we bold all approaches whose mean is not significantly less than the top-scoring mean IDA.

The first two datasets (Synthetic 1 and Synthetic 2) are binary datasets that demonstrate weaknesses in the Na\"ive approaches. We see that the Na\"ive methods match Base in IDA as they rejected no examples. In two-class datasets, the smallest maximum-softmax score is $0.5$. Hence, rejecting examples with the trivial $0.5$ threshold is unlikely as both softmax values must be exactly $0.5$. On the other hand, some B-CDF$_{\delta}$ approaches did reject confusing examples, with the B-CDF$_{.05/.10}$ variants yielding the highest IDA for Synthetic 1 and 2. In Synthetic 3, the B-CDF$_{.05/.10}$ variants produced the best results. Synthetic 4 again demonstrates the capability of B-CDF$_{\delta}$ to better model the ideal reject region, where all $\delta$ values gave statistically similar results above the other methods.

\newcommand{\imgwidth}{2cm}

\begin{table}[t]

    \caption{The mean/std ideal decision accuracy (IDA) of different approaches on four \textbf{equal} density synthetic datasets (R:B:G:Y) over 10 runs.}
    
    \setlength\tabcolsep{1.pt}
    
    \centering

    \begin{tabular}{c|c||c||c||c|}
    \cline{2-5}
    & Synthetic 1
    & Synthetic 2 
    & Synthetic 3
    & Synthetic 4\\
    \cline{2-5}
    
    &
    \includegraphics[width=\imgwidth]{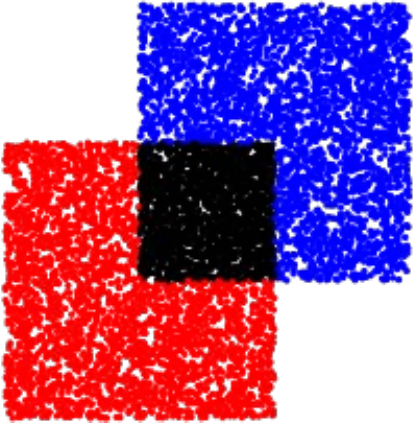} & \includegraphics[width=\imgwidth]{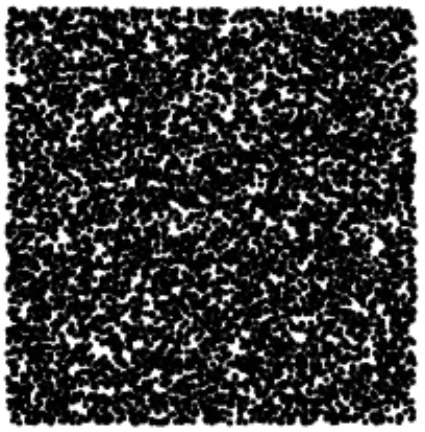} & \includegraphics[width=\imgwidth]{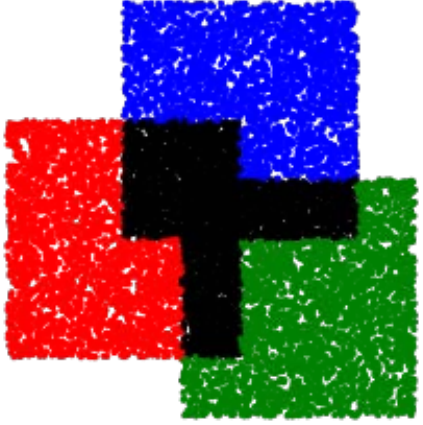} & \includegraphics[width=\imgwidth]{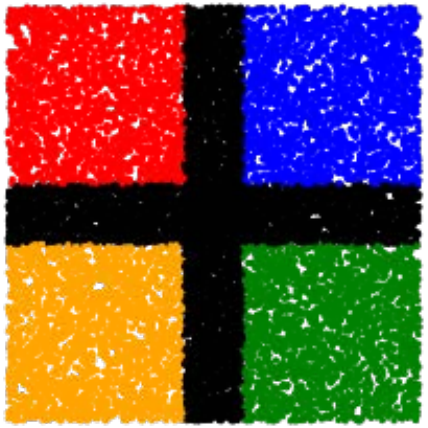}\\
  
    \hline
    Density Ratio & 1:1:0:0 
    & 1:1:0:0 
    & 1:1:1:0 
    & 1:1:1:1\\
    
    \hline
    \hline
    Method & IDA\boldmath{$\uparrow$}  & IDA\boldmath{$\uparrow$} & IDA\boldmath{$\uparrow$} &
     IDA\boldmath{$\uparrow$}\\
    \hline
    
    Base & 74.4  
    & 0.0 
    & 64.5 
    & 56.3\\
    \hline
    
    Na\"ive-NoCal & 74.4\textcolor{black!65}{\scriptsize$\pm$0.0} 
    & 0.0\textcolor{black!65}{\scriptsize$\pm$0.0} 
    & 70.4\textcolor{black!65}{\scriptsize$\pm$0.9}
    & 61.0\textcolor{black!65}{\scriptsize$\pm$0.3}\\

    Na\"ive-Cal & 74.4\textcolor{black!65}{\scriptsize$\pm$0.0}   
    & 0.0\textcolor{black!65}{\scriptsize$\pm$0.0} 
    & 71.6\textcolor{black!65}{\scriptsize$\pm$1.3} 
    & 64.5\textcolor{black!65}{\scriptsize$\pm$1.2} \\
    \hline
    
    B-CDF$_{.05}$ & \bfseries{76.7\textcolor{black!65}{\scriptsize$\pm$1.0} }
    & \bfseries{90.5\textcolor{black!65}{\scriptsize$\pm$29.7} }
    & \bfseries{88.4\textcolor{black!65}{\scriptsize$\pm$2.4} }
    & \bfseries{93.0\textcolor{black!65}{\scriptsize$\pm$1.3} }\\
    
    B-CDF$_{.10}$ & \bfseries{76.3\textcolor{black!65}{\scriptsize$\pm$0.9} }
    & \bfseries{70.8\textcolor{black!65}{\scriptsize$\pm$44.9}}
    & \bfseries{88.1\textcolor{black!65}{\scriptsize$\pm$2.2}}
    & \bfseries{93.2\textcolor{black!65}{\scriptsize$\pm$1.4}}\\
    
    B-CDF$_{.50}$ & 75.1\textcolor{black!65}{\scriptsize$\pm$1.6} 
    & 4.4\textcolor{black!65}{\scriptsize$\pm$6.0}
    & 85.4\textcolor{black!65}{\scriptsize$\pm$3.6}
    & \bfseries{93.7\textcolor{black!65}{\scriptsize$\pm$1.7}}\\
    
    B-CDF$_{.75}$ & 74.6\textcolor{black!65}{\scriptsize$\pm$0.1} 
    & 3.0\textcolor{black!65}{\scriptsize$\pm$6.3}
    & 80.5\textcolor{black!65}{\scriptsize$\pm$4.5}
    & \bfseries{93.7\textcolor{black!65}{\scriptsize$\pm$1.8} }\\
    
    B-CDF$_{.95}$ & 74.4\textcolor{black!65}{\scriptsize$\pm$0.0}
    & 0.0\textcolor{black!65}{\scriptsize$\pm$0.0}
    & 71.3\textcolor{black!65}{\scriptsize$\pm$4.4}
    & \bfseries{93.1\textcolor{black!65}{\scriptsize$\pm$2.0}}\\
    \hline

    \end{tabular} 
     
    \label{tab:synth_equal_density}
    \end{table}

\begin{table}[t]

    \caption{The mean select accuracy (SA), reject accuracy (RA), and coverage ($\phi$) of different approaches on four Gaussian \textbf{unequal} density synthetic datasets (R:B:G:Y) over 10 runs.}
    
    \setlength\tabcolsep{1.pt}
    
    \centering

    \begin{tabular}{c|cc|c||cc|c||cc|c||cc|c|}
    \cline{2-13}

    & \multicolumn{3}{c||}{Synthetic 5}
     & \multicolumn{3}{|c||}{Synthetic 6} & \multicolumn{3}{c||}{Synthetic 7} & \multicolumn{3}{|c|}{Synthetic 8}\\
    \cline{2-13}
    &
    \multicolumn{3}{|c||}{\includegraphics[width=\imgwidth]{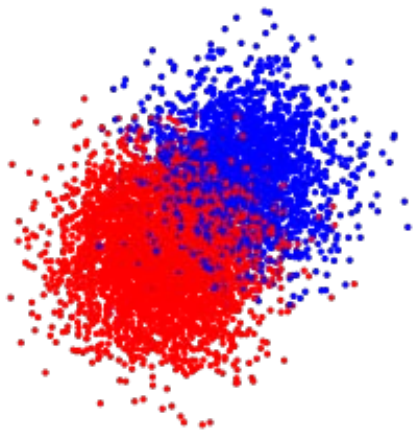}} & \multicolumn{3}{|c||}{\includegraphics[width=\imgwidth]{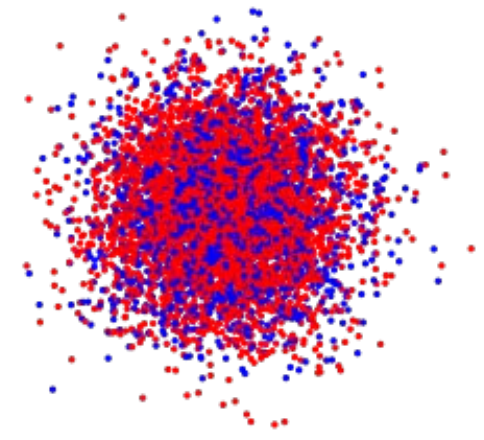}} & \multicolumn{3}{c||}{\includegraphics[width=\imgwidth]{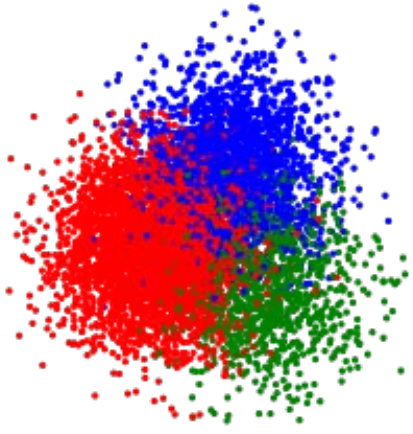}} & \multicolumn{3}{|c|}{\includegraphics[width=\imgwidth]{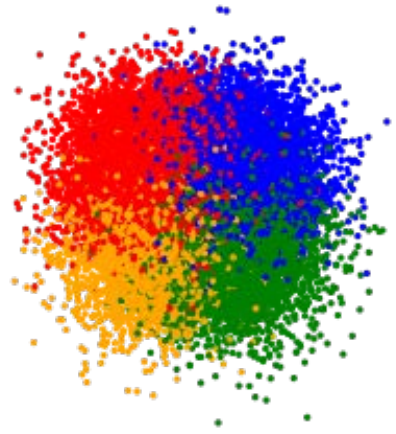}}\\
    \hline
    Density Ratio&
    \multicolumn{3}{|c||}{2:1:0:0} & \multicolumn{3}{|c||}{2:1:0:0} & \multicolumn{3}{c||}{4:2:1:0} & \multicolumn{3}{|c|}{6:5:4:3}\\
    \hline
    \hline
    Method & SA\boldmath{$\uparrow$} & RA\boldmath{$\downarrow$} &$\phi$\boldmath{$\uparrow$} & SA\boldmath{$\uparrow$} & RA\boldmath{$\downarrow$} &$\phi$\boldmath{$\uparrow$} & SA\boldmath{$\uparrow$} & RA\boldmath{$\downarrow$} &$\phi$\boldmath{$\uparrow$} & SA\boldmath{$\uparrow$} & RA\boldmath{$\downarrow$} &$\phi$\boldmath{$\uparrow$}\\
    \hline
    
    Base & 88.4 & -- & 100 
    & 66.6 & -- & 100 
    & 82.9 & -- & 100 
    & 71.4 & -- & 100\\
    \hline
    
    Na\"ive-NoCal & 88.4 & -- & 100 
    & 66.6 & -- & 100 
    &  84.2 & 42.7 & 96.3
    & 75.6 & 42.3 & 87.3\\

    Na\"ive-Cal & 88.4 & -- & 100 
    & 66.6 & -- & 100 
    & 84.8 & 42.3 & 94.9
    & 75.8 & 41.6 & 87.1\\
    \hline
    
    B-CDF$_{.05}$ &  91.2  &  54.3 &  92.6 
    & 66.7 & 61.5 & 95.5
    & 89.9 & 52.2 & 80.7
    & 83.0 & 52.2 & 62.4\\
    
    B-CDF$_{.10}$ &  90.5 & 52.8 & 94.5
    & 66.7 & 61.9 & 95.9
    & 89.5 & 51.2 & 82.2
    & 82.6 & 51.6 & 63.9\\
    
    B-CDF$_{.50}$ &  88.9 & 44.6 & 98.8
    & 66.7 & 53.8 & 98.4
    & 87.0 & 46.7 & 89.2
    & 81.0 & 49.7 & 69.3\\
    
    B-CDF$_{.75}$ &  88.6 & 38.7 & 99.6
    & 66.6 & -- & 100 
    & 85.6 & 44.1 & 92.9
    & 80.1 & 48.6 & 72.4\\
    
    B-CDF$_{.95}$ &  88.4 & -- & 100.0
    & 66.6 & -- & 100  
    & 84.1 & 39.3 & 96.8
    & 78.9 & 46.4 & 77.0\\
    \hline
    
    \end{tabular} 
     
    \label{tab:synth_gaussians}
    \end{table}
    
\subsubsection{Unequal Density Overlap}\hfill\\
Table \ref{tab:synth_gaussians} shows similarly configured datasets as Table \ref{tab:synth_equal_density} except that we used isotropic Gaussian distributions rather than uniform squares to sample the datasets. We also varied the density ratio of examples sampled per-class to evaluate regions of varying confusion. Here, the theoretical reject region is an infinitely thin line along the maximum posterior decision boundaries. However, this line could be a region \emph{in practice} due to the sampling of the datasets. Therefore, instead of employing IDA, we report and compare the select accuracy, reject accuracy, and coverage of the approaches.

We see in the 2-class datasets Synthetic 5 and 6 that the Base and Na\"ive methods have full coverage. However, the proposed algorithm identified viable reject regions in the sampled data. In Synthetic 5, the B-CDF$_{.05}$ variant improved the select accuracy by +2.8\% over the Base and Na\"ive methods, with a 54.3\% reject accuracy. In Synthetic 6, the B-CDF$_{.05}$ method scored a reject accuracy of 61.5\%, slightly larger than expected. Since thresholds are learned and ensured on validation data, this overage could occur on test data when using a strong model with a small reject region and different sampling of the data. Later in Sect. \ref{sec:generalize}, we will show a more detailed generalization experiment from validation to test data on real datasets. In Synthetic 7 and 8, we see that most B-CDF$_{\delta}$ approaches rejected more examples than Base and Na\"ive methods with reject accuracy near 50\%. The B-CDF$_{.05}$ variant improved the select accuracy over Base (+7\% and +11.6\%, respectively), while maintaining a viable reject accuracy of 52.2\%.

\subsection{Image Datasets}
We next evaluated the approaches on the benchmark image classification datasets CIFAR10 \cite{Cifar}, CIFAR100 \cite{Cifar}, FGVC-Aircraft \cite{Aircraft}, and ImageNet \cite{Imagenet} using pretrained state-of-the-art CNN and transformer models. These datasets contain various numbers of classes ranging from 10 to 1K. Since all reject approaches are post-processing methods, only the logits/softmax values of a trained model and the truth targets are needed. For CIFAR10, CIFAR100, and FGVC-Aircraft, we used pretrained state-of-the-art NAT CNNs \cite{cifarmodels}. For ImageNet, we used the pretrained BEiT large transformer \cite{imagenetmodel}. None of the aforementioned datasets have a fixed train-validation-test partitioning of the data. Therefore, we report scores on the validation set for ImageNet and the test set for CIFAR10, CIFAR100, and FGVC-Aircraft. Table \ref{tab:vision_results} shows the results of the various approaches.

Note that when comparing results from two different thresholds, if one yields higher select accuracy and lower reject accuracy (regardless of the coverage change), it is a definite improvement. If one has higher select accuracy and \emph{higher} reject accuracy, as long as the reject accuracy is within the acceptability level of the Binomial, then this is still considered an improvement.

\begin{table}[t]

    \caption{The select accuracy (SA), reject accuracy (RA), and coverage ($\phi$) of different approaches on four benchmark vision datasets.}
    
    \setlength\tabcolsep{1.pt}
    
    \centering

    \begin{tabular}{c|cc|c||cc|c||cc|c||cc|c|}
    \cline{2-13}

    & \multicolumn{3}{c||}{CIFAR10}
     & \multicolumn{3}{|c||}{CIFAR100} & \multicolumn{3}{|c||}{FGVC Aircraft} & \multicolumn{3}{|c|}{ImageNet}\\
\hline
    Method & SA\boldmath{$\uparrow$} & RA\boldmath{$\downarrow$} &$\phi$\boldmath{$\uparrow$} & SA\boldmath{$\uparrow$} & RA\boldmath{$\downarrow$} &$\phi$\boldmath{$\uparrow$} & SA\boldmath{$\uparrow$} & RA\boldmath{$\downarrow$} &$\phi$\boldmath{$\uparrow$} & SA\boldmath{$\uparrow$} & RA\boldmath{$\downarrow$} &$\phi$\boldmath{$\uparrow$}\\
    \hline
    
    Base & 98.4 & -- & 100 &
    88.3 & -- & 100 &
    90.1 & -- & 100 &
    88.4 & -- & 100\\
    \hline
    
    Na\"ive-NoCal & 98.5 & 57.1 & 99.9 &
    89.3 & 24.5 & 98.5 &
    96.3 & 58.4 & 83.5 &
    91.7 & 45.1 & 92.8\\

    Na\"ive-Cal & 98.6 & 48.8 & 99.6 &
    91.8 & 40.1 & 93.4 &
    93.2 & 33.3 & 94.8 &
    90.6 & 38.5 & 95.7\\
    \hline
    
    B-CDF$_{.05}$ &  99.3  & 58.7 & 97.9 &
     97.8  & 55.9 & 77.3 &
     98.3 & 45.4 & 84.4 &
     97.4  & 52.7 & 79.7\\
    
    B-CDF$_{.10}$ & 99.2 & 55.6 & 98.3 &
    97.3 & 53.1 & 79.7 &
    98.0 & 42.8 & 85.7 &
    96.9 & 49.5 & 81.9\\
    
    B-CDF$_{.50}$ & 98.9 & 42.5 & 99.2 &
    94.8 & 42.3 & 87.7 &
    96.2 & 28.6 & 91.0 &
    94.6 & 36.8 & 89.1\\
    
    B-CDF$_{.75}$ & 98.5 & 20.0 & 99.9 &
    92.3 & 32.6 & 93.4 &
    93.9 & 21.6 & 94.7 &
    92.1 & 27.7 & 94.1\\
    
    B-CDF$_{.95}$ & 98.4 & -- & 100 &
    89.3 & 21.3 & 98.6 &
    91.1 & 14.9 & 98.6 &
    89.7 & 22.5 & 98.0\\
    \hline
    
    \end{tabular} 
     
    \label{tab:vision_results}
    \end{table}

{\bfseries{CIFAR10}}. The Base approach scored $98.4\%$ select accuracy indicating that the model is already strong. Both Na\"ive approaches rejected a few examples, with Na\"ive-Cal scoring slightly higher select accuracy and lower reject accuracy than Na\"ive-NoCal. The highest select accuracy of $99.3\%$ was given by B-CDF$_{.05}$ with a coverage of $98.0\%$ and a reject accuracy of $58.7\%$. Though seemingly high, the B-CDF$_{\delta}$ approach statistically permits higher reject accuracy for smaller reject regions (consider the possible number of heads appearing on a small number of fair coin tosses). Comparing B-CDF$_{\delta}$ to Na\"ive-Cal, the B-CDF$_{.50}$ variant had a lower reject accuracy (42.5\%) and higher select accuracy (98.9\%), demonstrating better performance.

{\bfseries{CIFAR100}}. The Base approach scored $88.3\%$ select accuracy. The Na\"ive-Cal method rejected about $5\%$ more examples than Na\"ive-NoCal, yielding a higher select accuracy ($91.8\%$) and a lower reject accuracy (40.1\%). The B-CDF$_{.75}$ variant matched the coverage of Na\"ive-Cal while increasing select accuracy (+0.5\%) and decreasing reject accuracy (-7.5\%). These improvements indicate that the B-CDF$_{\delta}$ approach can better model confusion over the Na\"ive methods.

{\bfseries{FGVC-Aircraft}}. The Base approach scored $90.1\%$ select accuracy. The Na\"ive-Cal method outperformed Na\"ive-NoCal in coverage (94.8\%, 83.5\%, respectively) and reject accuracy (33.3\% and 58.4\%, respectively) but scored lower select accuracy (93.2\% and 96.3\%, respectively). Compared to Na\"ive-Cal, the B-CDF$_{.50/.75}$ variants scored much lower reject accuracy (-4.7\% and -11.7\%, respectively) and similar coverage (-3.8\% and -0.1\%, respectively) with increased select accuracy (+3.0\% and +0.7\%, respectively), indicating better performance.

{\bfseries{ImageNet}}. The Base approach scored $88.4\%$ select accuracy. The Na\"ive-NoCal and Na\"ive-Cal approaches achieved a higher select accuracy ($91.7\%$ and $90.6\%$, respectively) with reasonable reject accuracy ($45.1\%$ and $38.5\%$, respectively) and fairly high coverage (92.8\% and 95.7\%, respectively). Comparing the Na\"ive-Cal approach to B-CDF$_{\delta}$, the B-CDF$_{.50/.75}$ variants performed better in select accuracy (+4.0\% and +1.5\%, respectively) and reject accuracy (-1.7\% and -10.8\%, respectively). Nearly all B-CDF$_{\delta}$ approaches scored higher select accuracy and reasonable reject accuracy (as desired). We found that per-class thresholds varied widely on this dataset (and others with many classes).

\subsection{Text Datasets}
We next evaluated the approaches on the IMDB sentiment analysis \cite{sentiment} and AG News \cite{AG} text classification datasets. These datasets contain fewer classes than the image datasets (2 and 4, respectively). We utilized near state-of-the-art off-the-shelf pretrained BERT transformers \cite{Bert} for the evaluations. Both datasets do not include a validation set, hence we report results on the test data. Table \ref{tab:text_results} shows the results for text classification using the IMDB and AG News datasets.

{\bfseries{IMDB}}. The Base approach was fairly strong and scored a select accuracy of 94.7\%. Like in the 2-class synthetic datasets, the Na\"ive approaches failed to reject any examples. However, the B-CDF$_{\delta}$ method rejected some examples with B-CDF$_{.05}$ giving a reasonable reject accuracy (53.4\%) and greater select accuracy (+1.6\%). Other B-CDF$_{\delta}$ variants yielded higher select accuracy with reject accuracy near 50\%, except B-CDF$_{.95}$ which selected nearly all examples with 18.2\% reject accuracy and coverage of 99.9\%. 

{\bfseries{AG News}}. The Base method scored a select accuracy of 94.7\%. The Na\"ive-NoCal and Na\"ive-Cal approaches rejected only a few examples, improving select accuracy by +0.1\% and +0.2\%, respectively. The B-CDF$_{\delta}$ approach rejected more examples than Na\"ive-Cal, with B-CDF$_{.05}$ scoring the highest select accuracy (96.7\%) with acceptable reject accuracy (55.8\%). Moreover, all other B-CDF$_{\delta}$ variants (except B-CDF$_{.95}$) scored higher select accuracy than both Na\"ive methods, with a reasonable reject accuracy.

\begin{table}[t]

    \caption{The select accuracy (SA), reject accuracy (RA), and coverage ($\phi$) of different approaches on text datasets.}
    
    \setlength\tabcolsep{1.pt}
    
    \centering

    \begin{tabular}{c|cc|c||cc|c|}
    \cline{2-7}

    & \multicolumn{3}{c||}{IMDB}
     & \multicolumn{3}{c|}{AG News}\\
     
    \hline
    Method & SA\boldmath{$\uparrow$} & RA\boldmath{$\downarrow$} & $\phi$\boldmath{$\uparrow$} & SA\boldmath{$\uparrow$} & RA\boldmath{$\downarrow$} & $\phi$\boldmath{$\uparrow$} \\
    \hline
    
    Base & 94.7 & -- & 100
    & 94.7 & -- & 100 \\
    \hline
    
    Na\"ive-NoCal  & 94.7 & -- & 100 & 
    94.8 & 30.0 & 99.9\\

    Na\"ive-Cal & 94.7 & -- & 100 &
    94.9 & 33.3 & 99.8\\
    \hline
    
    B-CDF$_{.05}$ &  96.3  & 53.4 & 96.3 &
     96.7  & 55.8 & 95.1\\
    
    B-CDF$_{.10}$ & 96.2 & 52.7 & 96.5 &
    96.7 & 54.8 & 95.3\\
    
    B-CDF$_{.50}$ & 95.6 & 49.8 & 97.9 &
    
    95.6 & 46.0 & 98.2\\
    
    B-CDF$_{.75}$ &95.5 & 48.0 & 98.3  &
    
    95.4 & 39.8 & 98.8\\
    
    B-CDF$_{.95}$ & 94.7 & 18.2 & 99.9 &
    
    94.7 & -- & 100\\
    \hline

    \end{tabular} 
     
    \label{tab:text_results}
    \end{table}

\subsection{Generalization from Validation to Test Data}
\label{sec:generalize}
We now present results on how the learned reject thresholds generalize from a validation set to a test set. We employed CINIC10 \cite{CINIC10} (imagery) and Tweet Eval Emoji \cite{Tweet} (text) as both contain a proper train-validation-test partitioning. We utilized a weakly-trained ResNet20 CNN \cite{resnet} for CINIC10 and an off-the-shelf pretrained BERT transformer \cite{Bert} for Tweet Eval Emoji. Table \ref{tab:transfer_results} shows the results of rejection thresholds learned from validation and applied to test data. 

CINIC10 contains 90K training, 90K validation, and 90K testing examples. Given that the validation set here is large (equal to the test set), we expect thresholds to behave similarly across validation and test. We see that select accuracy, reject accuracy, and coverage all transfer well to the test set (nearly a one-to-one match). The earlier Figs. \ref{fig:cinic10_tsne_select} and \ref{fig:cinic10_tsne_reject} depict t-SNE embeddings computed using the select and reject sets given by B-CDF$_{.05}$.   

Tweet Eval Emoji contains 45K training, 5K validation, and 50K testing examples with 21 classes. For this dataset, the validation set is 10X \emph{smaller} than the test set. Although, we observe that all metrics here tend to be higher (desired) on the test set across all approaches, which shows that the proposed algorithm compensates in the proper direction. However, in Table \ref{tab:synth_gaussians} the reject accuracy of the B-CDF$_{\delta}$ approach on Synthetic 6 was larger than expected due to the smaller validation set. We additionally examined a larger validation set on Synthetic 6 and saw expected generalization performance.

\begin{table}[t]

    \caption{The generalization of select accuracy (SA), reject accuracy (RA), and coverage ($\phi$) of thresholds learned on validation and applied to test data for CINIC10 and Tweet Eval Emoji.}
    
    \setlength\tabcolsep{1.pt}
    
    \centering

    \begin{tabular}{c|cc|c||cc|c||cc|c||cc|c|}
    \cline{2-13}

    & \multicolumn{6}{c||}{CINIC10}
     & \multicolumn{6}{c|}{Tweet Eval Emoji} \\
     
    \cline{2-13}

    & \multicolumn{3}{|c||}{Val}
     & \multicolumn{3}{|c||}{Test} 
     & \multicolumn{3}{|c||}{Val} 
     & \multicolumn{3}{|c|}{Test}\\
\hline
    Method & SA\boldmath{$\uparrow$} & RA\boldmath{$\downarrow$} &$\phi$\boldmath{$\uparrow$} & SA\boldmath{$\uparrow$} & RA\boldmath{$\downarrow$} &$\phi$\boldmath{$\uparrow$} & SA\boldmath{$\uparrow$} & RA\boldmath{$\downarrow$} &$\phi$\boldmath{$\uparrow$} & SA\boldmath{$\uparrow$} & RA\boldmath{$\downarrow$} &$\phi$\boldmath{$\uparrow$}\\
    \hline
    
    Base & 81.8 & -- & 100 &
    81.5 & -- & 100 & 
    32.6 & -- & 100 &
    47.9 & -- & 100\\
    \hline
    
    Na\"ive-NoCal & 83.5 & 31.4 & 96.9 &
    83.1 & 29.7 & 96.9 & 
    58.5 & 22.5 & 28.0 &
    74.7 & 24.7 & 46.5\\

    Na\"ive-Cal & 87.7 & 40.2 & 87.7 &
    87.5 & 39.3 & 87.4 & 
    69.3 & 24.0 & 19.1 &
    79.4 & 28.5 & 38.2\\
    \hline
    
    B-CDF$_{.05}$ &  93.4  & 51.6 & 72.5 &
     93.2   & 51.0 & 72.2 & 
     92.9 & 30.3 & 3.6 &
    93.7  & 35.3 & 21.7\\
    
    B-CDF$_{.10}$ & 93.2 & 51.2 & 73.0 &
    93.0 & 50.6 & 72.8 & 
    90.7 & 30.1 & 4.1 &
    92.4 & 35.0 & 22.5\\
    
    B-CDF$_{.50}$ & 92.6 & 49.9 & 74.8 &
    92.4 & 49.3 & 74.6 & 
    82.8 & 29.2 & 6.4 &
    89.1 & 33.8 & 25.6\\
    
    B-CDF$_{.75}$ & 92.1 & 49.2 & 76.1 &
    92.0 & 48.4 & 75.8 & 
    79.7 & 28.4 & 8.2 &
    86.6 & 33.1 & 27.7\\
    
    B-CDF$_{.95}$ & 91.5 & 48.1 & 77.7 &
    91.4 & 47.3 & 77.4 & 
    71.3 & 27.4 & 11.9 &
    82.0 & 32.8 & 30.7\\
    \hline
    
    \end{tabular} 
     
    \label{tab:transfer_results}
    \end{table}

\subsection{Alternative Confidence Interval Formulations}
Given the computational complexity of evaluating the Binomial CDF on very large reject regions, we provide alternative confidence interval methods with a similar goal but lower computational complexity. The objective remains the same as B-CDF$_\delta$, but the randomness evaluation uses a one-sided confidence interval to determine whether the upper bound on \emph{true} accuracy given the \emph{observed} accuracy is reasonable (less than or equal to 50\%). We present results using the Clopper-Pearson interval \cite{Clopper1934}, the Wilson interval with and without continuity correction (Wilson-CC and Wilson-NoCC, respectively) \cite{Brown2001}, and the Agresti Coull interval \cite{Agresti1998}. Closed-form equations exist for Wilson-CC, Wilson-NoCC, and Agresti Coull. Table \ref{tab:alternative_results} shows the results using these confidence intervals.

\begin{table}[t]

    \caption{The select accuracy (SA), reject accuracy (RA), and coverage ($\phi$) of alternative confidence interval approaches on four benchmark vision datasets.}
    
    \setlength\tabcolsep{1.pt}
    
    \centering

    \begin{tabular}{c|cc|c||cc|c||cc|c||cc|c|}
    \cline{2-13}

    & \multicolumn{3}{c||}{CIFAR10}
     & \multicolumn{3}{|c||}{CIFAR100} & \multicolumn{3}{|c||}{FGVC Aircraft} & \multicolumn{3}{|c|}{ImageNet}\\
\hline
    Method & SA \boldmath{$\uparrow$} & RA \boldmath{$\downarrow$} &$\phi$\boldmath{$\uparrow$} & SA \boldmath{$\uparrow$} & RA \boldmath{$\downarrow$} &$\phi$\boldmath{$\uparrow$} & SA \boldmath{$\uparrow$} & RA \boldmath{$\downarrow$} &$\phi$\boldmath{$\uparrow$} & SA \boldmath{$\uparrow$} & RA \boldmath{$\downarrow$} &$\phi$\boldmath{$\uparrow$}\\
    \hline
    
    Base & 98.4 & -- & 100 &
    88.3 & -- & 100 &
    90.1 & -- & 100 &
    88.4 & -- & 100\\
    \hline
    
    Na\"ive-NoCal & 98.5 & 57.1 & 99.9 &
    89.3 & 24.5 & 98.5 &
    96.3 & 58.4 & 83.5 &
    91.7 & 45.1 & 92.8\\

    Na\"ive-Cal & 98.6 & 48.8 & 99.6 &
    91.8 & 40.1 & 93.4 &
    93.2 & 33.3 & 94.8 &
    90.6 & 38.5 & 95.7\\
    \hline
    
    B-CDF$_{.05}$ &  99.3  & 58.7 & 97.9 &
    97.8 & 55.9 & 77.3 &
    98.3 & 45.4 & 84.4 &
    97.4 & 52.7 & 79.7\\
    \hline
    
    Clopper-Pearson &  99.3  & 58.7 & 97.9 &
    97.8 & 55.9 & 77.3 &
    98.3 & 45.4 & 84.4 &
    97.4 & 52.7 & 79.7\\
    
    Wilson-CC &  99.3 & 58.7 & 97.9 &
    97.8 & 55.9 & 77.3 &
    98.3 & 45.4 & 84.4 &
    97.4 & 52.7 & 79.7\\
    
    Wilson-NoCC &  99.3 & 59.8 & 97.7 &
     98.2  & 57.8 & 75.6 &
    98.7 & 49.1 & 82.5 &
    97.7 & 54.9 & 78.1\\
    
    Agresti-Coull &  99.3 & 59.8 & 97.7 &
     98.2 & 57.8 & 75.6 &
     98.8 & 49.7 & 82.3 &
     97.8 & 55.0 & 78.0\\
    \hline
    
    \end{tabular} 
     
    \label{tab:alternative_results}
    \end{table}

On these datasets, the Clopper-Pearson and Wilson-CC approaches arrive at the same solutions as the original B-CDF$_{.05}$ approach. However, Wilson-NoCC and Agresti-Coull typically score higher select accuracy at the cost of higher reject accuracy and lower coverage. The Agresti-Coull method scored the highest select accuracy, highest reject accuracy, and lowest coverage for all experiments. We showed results based on $\delta\!\!=\!\!.05$, but for all significance levels examined in this paper, we found that Clopper-Pearson and Wilson-CC approaches matched B-CDF$_{\delta}$. These results demonstrate that Clopper-Pearson and Wilson-CC could be interchanged with B-CDF$_{\delta}$ to reduce computational complexity, if desired.

\subsection{Discussion}
We evaluated multiple approaches on equal-density synthetic datasets and showed that the B-CDF$_\delta$ method provides the highest accuracy to the ideal decision function. On unequal-density synthetic data, real-world imagery, and text datasets, our approach performed the best at specific $\delta$ values, yielding the highest select accuracy while keeping a reasonable and statistically viable reject accuracy (near 50\%). Furthermore, given a user preference for select accuracy or coverage in a specific application, the user could examine different $\delta$ values to best suit the task. Higher values of $\delta$ provide increased coverage, while lower values of $\delta$ provide increased select accuracy. Overall, we found that lower values of $\delta$ seem preferable across multiple datasets and could be used as a default. We have also shown that thresholds learned on large validation sets transfer well to test sets. Lastly, we presented an alternative formulation using related confidence intervals and showed that they could provide similar performance at reduced computational complexity. Therefore, when given very large datasets with extensive reject regions, it is recommended to use alternative formulations.

\section{Conclusion}
Given the growing adoption of neural networks in various applied tasks, the need for confident predictions is becoming increasingly important. We proposed a Binomial-CDF approach to automatically detect and filter out regions of confusion for any neural classifier and dataset pairing. This post-processing technique leverages a validation set to learn per-class rejection thresholds that can identify and reject regions in the decision space based on random-chance classification. This approach is applicable when strong constraints on select accuracy or coverage are unavailable. We demonstrated that the approach provides a favorable scoring of select and reject accuracy on 2-D points, imagery, and text datasets. In future work, we plan to develop a joint-training objective to learn the rejection function during the training of the neural network.\\

\noindent {\bfseries Acknowledgements.} This research was supported by the U.S. Air Force Research Laboratory under Contract \#GRT00054740 (Release \#AFRL-2022-3339).

%
%
%
\bibliographystyle{splncs04}
\bibliography{egbib}

\end{document}